\documentclass[11pt]{article}

\usepackage[final]{acl}

\usepackage{times}
\usepackage{latexsym}
\usepackage[T1]{fontenc}
\usepackage[utf8]{inputenc}
\usepackage{microtype}
\usepackage{inconsolata}
\usepackage{graphicx}
\usepackage{url}
\usepackage{booktabs}
\usepackage{listings}
\usepackage{xcolor}
\usepackage{enumitem}
\usepackage{amsmath}
\usepackage{amssymb}
\usepackage{adjustbox}
\usepackage{array}
\usepackage{fancyvrb}
\usepackage{fvextra}
\usepackage{tikz}
\usepackage{tabularx}
\usetikzlibrary{arrows.meta,positioning,fit,calc}

\definecolor{darkblue}{rgb}{0, 0, 0.5}
\definecolor{codegray}{rgb}{0.96,0.96,0.96}
\definecolor{codecomment}{rgb}{0.13,0.45,0.13}
\definecolor{codekeyword}{rgb}{0.10,0.20,0.65}
\definecolor{codestring}{rgb}{0.55,0.10,0.10}
\hypersetup{colorlinks=true, citecolor=darkblue, linkcolor=darkblue, urlcolor=darkblue}

\lstdefinestyle{acljson}{%
  basicstyle=\ttfamily\tiny,
  breaklines=true,
  breakatwhitespace=false,
  breakindent=0pt,
  columns=fullflexible,
  linewidth=0.96\linewidth,
  backgroundcolor=\color{codegray},
  frame=single,
  rulecolor=\color{black!20},
  framesep=3pt,
  framerule=0.4pt,
  xleftmargin=0pt,
  xrightmargin=0pt,
  aboveskip=6pt,
  belowskip=6pt
}

\title{Extending FunctionGemma for Practical On-Device Mobile Function Calling}

\author{Ali Rezagholizadeh \\
  { } \\
  \texttt{alirezagholizadeh70@gmail.com} \\
  \And
  Soheila Samiee \\
  { } \\
  \texttt{sla.samiee@gmail.com} \\}

\newcommand{\datasetExtended}{\textsc{Mobile\-Actions\-Extended}}
\newcommand{\datasetGoogle}{\textsc{Mobile\-Actions\-Google}}
\newcommand{\datasetCombined}{\textsc{Mobile\-Actions\-Combined}}
\newcommand{\modelBase}{\textsc{FunctionGemma\-270M\-it Base}}
\newcommand{\modelGoogleFT}{\textsc{FunctionGemma-270M-Mobile-Actions}}
\newcommand{\modelOurs}{\textsc{FunctionGemma-270M-it-Mobile-Actions-Extended}}
\newcommand{\modelCombined}{\textsc{FunctionGemma-270M-it-Mobile-Actions-Combined}}

\newcommand{\Mbase}{Base}
\newcommand{\Mgft}{Google~MA}
\newcommand{\MoursE}{Ours: MA-Extended}
\newcommand{\MoursC}{Ours: MA-Combined}

\newcommand{\bk}{\discretionary{}{}{}}

\newcommand{\dataExtRepo}{\texttt{UGrowAI/Mobile-Actions-Extended}}
\newcommand{\dataMergeRepo}{\texttt{UGrowAI/Mobile-Actions-Combined}}

\newcommand{\modelExtRepo}{\texttt{UGrowAI/FunctionGemma-270M-it-Mobile\newline-Actions-Extended}}
\newcommand{\modelCombRepo}{\texttt{UGrowAI/FunctionGemma-270M-it-Mobile\newline-Actions-Combined}}

\newcommand{\modelExtLitRepo}{\texttt{UGrowAI/FunctionGemma-270M-it-Mobile\newline-Actions-Extended-litert-lm}}
\newcommand{\modelCombLitRepo}{\texttt{UGrowAI/FunctionGemma-270M-it-Mobile\newline-Actions-Combined-litert-lm}}

\newcommand{\modelOursLitRepo}{\footnote{https://huggingface.co/UGrowAI/FunctionGemma-270M-it-Mobile-Actions-Extended-litert-lm}}

\newif\ifhighlightrevisions
\highlightrevisionstrue
\ifhighlightrevisions

\else

\fi

\DefineVerbatimEnvironment{demotrace}{Verbatim}{%
  fontsize=\small,
  frame=single,
  framesep=3pt,
  rulecolor=\color{black!20},
  breaklines=true,
  breakanywhere=true,
  breaksymbolleft={},
  breaksymbolright={}}

\begin{document}

\maketitle

\begin{abstract}
On-device assistants require function-calling models that map natural language to local system actions, but existing resources emphasize web APIs or narrow mobile-action catalogs. We extend FunctionGemma~270M-it to practical Android workflows by introducing \datasetExtended{}, a synthetic, schema-validated dataset of $\sim$9{,}500 conversations covering fifteen device-control categories, including messaging, phone calls, camera/screenshot, brightness control, device-status queries, flashlight control, and application management. We fine-tune the 270M model with TRL supervised fine-tuning under completion-only loss, producing an extended specialist and a combined model trained jointly with Google's \datasetGoogle{}. On \datasetExtended{}, end-to-end accuracy improves from $29.3\%$ for the base model and $17.2\%$ for Google's Mobile-Actions variant to $76.5\%$. The combined model retains $76.5\%$ on \datasetExtended{} and reaches $82.3\%$ on \datasetGoogle{}, down from the $90.3\%$ of Google's Mobile-Actions specialist, representing an 8.0-percentage-point trade-off in return for doubling category coverage. 
We release the \href{https://huggingface.co/UGrowAI/datasets}{dataset},
fine-tuned \href{https://huggingface.co/UGrowAI/models}{models}
, reproducible training/evaluation pipeline
, and an \href{https://github.com/UgrowAI/HandyChat-apk/releases/latest/download/HandyChat.apk}{Android demo}, highlighting compact local function calling as a practical path towards low-latency and privacy-preserving mobile assistants.

\end{abstract}

\section{Introduction}
\label{sec:intro}

Modern conversational assistants increasingly bridge natural language to executable actions through \emph{function calling}: instead of generating free-form text, the model emits a structured tool call (a name and a JSON-like dictionary of arguments) that an external runtime can dispatch to operating-system APIs, web services, or local hardware \citep{schick2023toolformer,yao2022react,patil2024gorilla,openai-function-calling-2023,anthropic-tool-use-2024}. The function-calling formulation is now standard in commercial APIs and underlies most agent frameworks \citep{wang2026function}.

While the formulation is general, deployments differ sharply. Server-side agents primarily call \emph{web APIs} (REST endpoints, search, calendars, SaaS tools), and most public benchmarks reflect this bias \citep{patil2024gorilla,berkeley-fcl-2024,liu2024apigen,zhuo2024bigcodebench}. On-device assistants, by contrast, rely on \emph{system actions} (e.g., turning the flashlight on, sending an SMS, opening an installed application, querying the battery) that must be invoked through the host operating system rather than the network. These actions are high-frequency, latency-sensitive, and privacy-sensitive.

This shift creates demand for \emph{compact} function-calling models that can run on a phone. Sub-billion-parameter models such as FunctionGemma~270M-it~\citep{functiongemma-2025}, built on the Gemma~3 family~\citep{gemma2024,gemma3-270m-2025}, occupy this niche. However, their training distribution still skews toward general tool use; the ready-made on-device specialization, FunctionGemma~270M Mobile Actions \citep{functiongemma-mobile-actions-2025}, focuses on roughly seven mobile workflows (calendar event creation, contact creation, mapping, email, Wi-Fi settings, flashlight on/off), leaving common everyday intents, like phone calls, SMS, brightness control, camera, screenshots, and application opening uncovered.

In this study, we evaluate whether a small, targeted dataset of mobile-action conversations is sufficient to extend a 270M-parameter function-calling model to a meaningfully broader set of on-device intents \emph{without} catastrophically forgetting the original distribution. Concretely, we investigate three questions:

\begin{itemize}[leftmargin=1.2em,itemsep=2pt,topsep=2pt]
    \item \textbf{(Q1) Coverage gap.} How well does the publicly available FunctionGemma model handle mobile actions outside of its training distribution?
    \item \textbf{(Q2) Targeted fine-tuning.} Does full supervised fine-tuning on a synthetically generated, schema-validated dataset of 15 mobile categories close that gap on a 270M backbone?
    \item \textbf{(Q3) Cross-domain transfer.} Can a single fine-tuned model serve both the original (Google Mobile Actions) and the new (device-action) distributions?
\end{itemize}

Our contributions can be summarized as:

    (i) \textbf{An open mobile-actions dataset.} We release \datasetExtended{}, a $\sim$9{,}500-example function-calling corpus generated through schema-grounded synthetic prompting, post-validated for tool-call legality, and split 90/10 between train and evaluation (Section~\ref{sec:dataset}).
    
    (ii) \textbf{Two extended FunctionGemma fine-tuned models.} We fine-tune FunctionGemma~270M-it with Transformers Reinforcement Learning (TRL) supervised fine-tuning \citep{trl,wolf2020transformers} under completion-only loss, releasing \modelOurs{} and \modelCombined{}.
    
    (iii) \textbf{A multi-axis evaluation.} We compare four versions of the model on two held-out splits using exact-match metrics for function names, arguments, and end-to-end correctness (Section~\ref{sec:experiments}).
    
    (iv) \textbf{An end-to-end pipeline.} We open-source the full reproduction pipeline---from dataset generation through Hugging Face publishing, TRL fine-tuning, and evaluation---as three independent packages.
    
    (v) \textbf{An on-device demo app (\emph{Handy Chat}).} We deploy .litertlm version of \modelOurs{} in a Buildozer-packaged Android app (Python Kivy UI, PyJNIus Java bridge to \texttt{litertlm-android}, LiteRT-LM inference on an exported \texttt{.litertlm} bundle) because the Google AI Edge Gallery~\citep{ai-edge-gallery} does not support custom checkpoints in its Mobile Actions workflow (Sections~\ref{sec:demo}, Appendix~\ref{app:demo-section}).

Our extended model lifts end-to-end accuracy on \datasetExtended{} from $29.3\%$ (base) and $17.2\%$ (Google's Mobile Actions variant) to $76.5\%$, with perfect function-name accuracy on all fifteen categories. The \emph{combined} model preserves competitive performance on \datasetGoogle{}: end-to-end accuracy drops from $90.3\%$ for the Google specialist to $82.3\%$, an 8.0-percentage-point cost that we consider an acceptable trade-off for doubling category coverage. Remaining errors concentrate on optional-boolean formatting (\texttt{dial\_call}, \texttt{create\_chooser}). Per-category breakdown and failure analysis are presented in Appendix~\ref{app:percat}.

\section{Related Work}
\label{sec:related}

Modern tool-using LLMs~\citep{schick2023toolformer,yao2022react,patil2024gorilla,qin2024toolllm,openai-function-calling-2023} replace classical intent-and-slot pipelines with schema-conditioned structured output. Compact function-calling models, such as Granite Function Calling~\citep{abdelaziz2024granite}, Hammer~\citep{lin2024hammer}, ToolACE~\citep{liu2025toolace}, and TinyAgent~\citep{erdogan2024tinyagent} show that targeted fine-tuning, not scale alone, drives quality at the edge \citep{singh2025smallmodelsbigtasks}. FunctionGemma~270M-it~\citep{functiongemma-2025} and Google's Mobile Actions variant~\citep{functiongemma-mobile-actions-2025} target on-device deployment; however, available benchmarks such as BFCL~\citep{berkeley-fcl-2024}, BigCodeBench~\citep{zhuo2024bigcodebench}, and APIGen~\citep{liu2024apigen} focus on web APIs rather than Android system actions. To the best of our knowledge, \datasetGoogle{} is the only public mobile-actions function-calling benchmark, and this study aims to extend it. 
A more detailed summary of related works is available in Appendix~\ref{app:related}.

\section{Task Definition and Tool Catalog}
\label{sec:task}

\begin{table*}[t]
\centering
\small
\begin{tabularx}{\textwidth}{c p{0.21\textwidth} p{0.3\textwidth} X}
\hline
\textbf{Count} & \textbf{Function} & \textbf{Description} & \textbf{Arguments} \\
\hline
1  & \texttt{turnOn\_light}            & Turn the flashlight on.             & None \\
2  & \texttt{turnOff\_light}           & Turn the flashlight off.            & None \\
3  & \texttt{send\_email}              & Send an email.                      & \texttt{to}, \texttt{subject}, optional \texttt{body}, optional \texttt{create\_chooser} \\
4  & \texttt{battery\_status}          & Query battery level/state.          & None \\
5  & \texttt{bluetooth\_status}        & Query Bluetooth state.              & None \\
6  & \texttt{phone\_call}              & Place a call.                       & \texttt{number}, \texttt{dial\_call}: dial-pad vs.\ direct \\
7  & \texttt{phone\_sms}               & Send an SMS message.                & \texttt{sms\_recipient}, optional \texttt{sms\_message} \\
8  & \texttt{take\_picture}            & Take a photo via the camera.        & None \\
9  & \texttt{take\_screenshot}         & Capture the current screen.         & None \\
10 & \texttt{get\_current\_brightness} & Read the current screen brightness. & None \\
11 & \texttt{decrease\_brightness}     & Step the brightness down.           & None \\
12 & \texttt{increase\_brightness}     & Step the brightness up.             & None \\
13 & \texttt{set\_brightness}          & Set brightness to a specific level. & \texttt{level} in $[0,100]$ \\
14 & \texttt{list\_application}        & List installed applications.        & None \\
15 & \texttt{open\_application}        & Open an application by name.        & \texttt{application\_name} \\
\hline
\end{tabularx}

\caption{Summary of the new Mobile-Action categories: supported functions, descriptions, and arguments.}
\label{tab:function_summary}
\end{table*}

\subsection{Function Calling as Structured Output}

We use the standard formulation. A \emph{tool catalog} $\mathcal{T} = \{T_1, \ldots, T_K\}$ is a set of typed function declarations; each $T_k$ has a name, a natural-language description, and a parameter schema. Given a system prompt, the catalog $\mathcal{T}$, and a user utterance $u$, the model must produce an ordered sequence of calls
\begin{equation*}
\begin{split}
\hat{c}_1, \hat{c}_2, \ldots, \hat{c}_m,\ \hat{c}_i &= \langle \text{name}_i, \text{args}_i \rangle, \\
&\text{name}_i \in \{T_1, \ldots, T_K\}.
\end{split}
\end{equation*}
\noindent FunctionGemma renders calls and declarations with sentinel tokens (\texttt{<start\_function\_call>}, \texttt{<end\_function\_call>}, \texttt{<escape>}) so that output can be deterministically parsed.

\subsection{Evaluation Granularity}

Following \citet{berkeley-fcl-2024}, the evaluation is executed at three levels: \textbf{function-name accuracy} (exact match on ordered name lists), \textbf{argument exact match} (exact match on key-sorted argument dictionaries), and \textbf{end-to-end correctness} (logical AND). This strict criterion flags any extra or missing key as failure; we therefore complement it with an \emph{execution-aware} variant (Section~\ref{sec:exec-aware}) and discuss implications in Section~\ref{sec:discussion}.

\subsection{New Mobile-Action Categories}
\label{sec:categories}

In this study, we extend FunctionGemma's mobile coverage with the fifteen tools presented in Table~\ref{tab:function_summary}. The schemas (full JSON in Appendix~\ref{app:schema}) mirror Plyer~\citep{plyer}-style cross-platform Android APIs so that every call can be dispatched to a real handler in a demo runtime.
These categories cover daily ``physical'' assistant intents absent from \datasetGoogle{} (calls, SMS, brightness, camera, screenshots, app management) and are implementable through Plyer on Android.

\section{Dataset Construction: \datasetExtended{}}
\label{sec:dataset}

\begin{figure*}[t]
\centering
\resizebox{0.8\textwidth}{!}{%
\begin{tikzpicture}[
  node distance=0.24cm and 0.28cm,
  box/.style={draw=black!70, rounded corners=2pt, align=center,
    font=\scriptsize, inner sep=2.5pt, minimum height=0.68cm, fill=white},
  databox/.style={box, fill=black!4},
  enddatabox/.style={box, fill=black!12},
  trainbox/.style={box, fill=black!8},
  exportbox/.style={box, fill=black!6},
  demobox/.style={box, fill=black!12, font=\scriptsize\bfseries},
  arr/.style={-{Stealth[length=1.8mm]}, semithick},
  lbl/.style={font=\tiny, align=center}
]

\node[databox, minimum width=1.35cm] (schema) {Prompt};
\node[databox, right=of schema, minimum width=1.25cm] (teacher) {Teacher\\LLM};
\node[databox, right=of teacher, minimum width=1.35cm] (raw) {Synthetic\\pairs};
\node[databox, right=of raw, minimum width=1.2cm] (valid) {Schema\\validator};
\node[databox, right=of valid, minimum width=1.25cm] (complete) {FG chat\\completer};
\node[enddatabox, right=of complete, minimum width=1.35cm] (ext) {MA-Ext.\\{\tiny $\sim$9.5K}};

\draw[arr] (schema) -- (teacher);
\draw[arr] (teacher) -- (raw);
\draw[arr] (raw) -- (valid);
\draw[arr] (valid) -- (complete);
\draw[arr] (complete) -- (ext);

\node[databox, below=0.40cm of complete, minimum width=1.2cm] (google) {MA-\\Google};
\node[enddatabox, below=0.40cm of ext, minimum width=1.35cm] (merge) {MA-\\Combined};

\draw[arr] (google) -- (merge);
\draw[arr] (ext.south) -- (merge.north);

\node[trainbox, below=1.5cm of schema, minimum width=1.45cm] (row2merge) {MA-\\Combined};
\node[trainbox, below=0.40cm of row2merge, minimum width=1.35cm] (row2ext) {MA-Ext.\\{\tiny $\sim$9.5K}};
\node[trainbox, below= 1.7cm of teacher, right=0.42cm of row2merge, right=0.42cm of row2merge, minimum width=1.45cm] (sft) {TRL\\full SFT};
\node[trainbox, below= 1.5cm of raw, right=0.45cm of sft, minimum width=1.35cm] (oursCE) {Ours–C\\Ours–E};
\node[trainbox, below=1.5cm of valid, right=0.42cm of oursCE, minimum width=1.25cm] (eval) {Exact-match\\Evaluation};
\node[trainbox, below=0.4cm of oursCE, minimum width=1.25cm] (litert) {Litertlm};
\node[demobox, below=1.7cm of merge, right=0.42cm of litert, minimum width=2.15cm] (handy) {Handy Chat\\{\tiny Kivy + Buildozer}\\{\tiny Java bridge + LiteRT-LM}};

\draw[arr] (row2merge) -- (sft);
\draw[arr] (row2ext) -- (sft);
\draw[arr] (sft) -- (oursCE);
\draw[arr] (oursCE) -- (eval);
\draw[arr] (oursCE.south) -- (litert.north);
\draw[arr] (litert) -- (handy);

\node[lbl, above=0.05cm of teacher] {data generation};
\node[lbl, above=0.05cm of oursCE] {fine-tuned\\model};
\node[lbl, left=0.05cm of litert] {lightweight\\model};
\end{tikzpicture}%
}
\caption{End-to-end pipeline. \textbf{Top:} schema-grounded synthetic generation, validation, FunctionGemma chat completion, and merge with Google's mobile-actions corpus. \textbf{Bottom:} serving generated datasets, TRL fine-tuning, fine-tuned model, exact-match evaluation, export lightweight \texttt{.litertlm} model quantized by 8-bit integer weight and 32-bit float activation, and typed-input deployment in \emph{Handy Chat} (Buildozer APK, PyJNIus Java bridge to \texttt{litertlm-android}, Plyer dispatch). Dynamic INT8 quantization (\texttt{dynamic\_wi8\_afp32}: 8-bit integer weights, 32-bit float activations) is applied \emph{solely} to the exported \texttt{.litertlm} bundle used by the deployed LiteRT-LM Android demo; all evaluation accuracy figures reported in this paper were measured on the unquantized Hugging Face checkpoints.
(more details in Sections
~\ref{sec:dataset}--\ref{sec:demo} and
Appendices~\ref{app:pipeline}
and ~\ref{app:demo-section}.)}
\label{fig:pipeline}
\end{figure*}
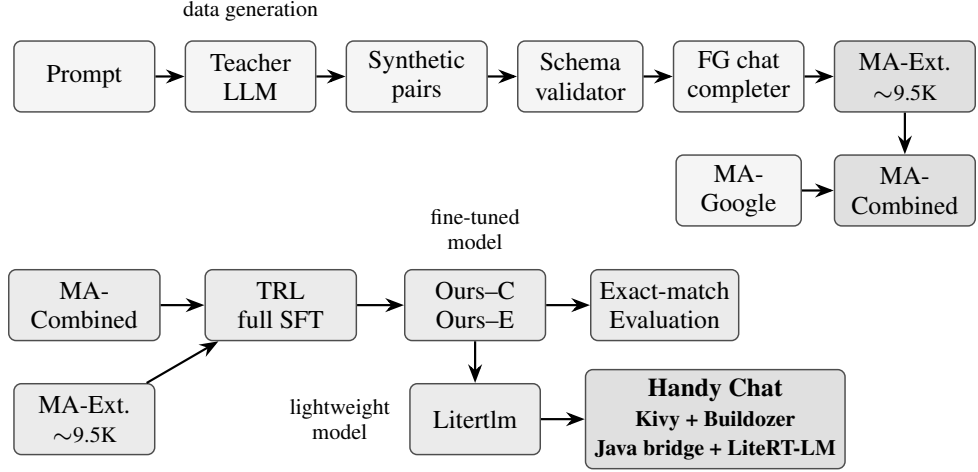

We aimed for a dataset that (i) is large enough to fine-tune a 270M-parameter backbone end to end, (ii) follows FunctionGemma's exact chat-template format, and (iii) is formatted similar to \datasetGoogle{}~\citep{google-mobile-actions-dataset-2025} so the two datasets can be merged easily for cross-domain training. Figure~\ref{fig:pipeline} summarizes the full workflow from schema-grounded generation through fine-tuning, evaluation, and on-device deployment.

\subsection{Schema-Grounded Synthetic Generation}

Raw conversations are generated by prompting a strong teacher with the JSON tool list and a worked output example. The teacher produces (i) $\sim$300 user prompts per single tool and (ii) $\sim$5{,}000 multi-tool prompts (e.g., \emph{``check battery and take a screenshot''}). Each record is a \texttt{user}/\texttt{assistant} pair with gold \texttt{tool\_calls}. The full prompt is in Appendix~\ref{app:gen-prompt}. This recipe follows APIGen~\citep{liu2024apigen} and ToolACE~\citep{liu2025toolace}; unlike APIGen we do not execute calls during generation, relying instead on structural validation.

\subsection{Validation and Completion}

A validator (\texttt{validate\_generated\_dataset.py}) parses every assistant message and checks that: (i)
    every called function name is present in the catalog;
    (ii) all \texttt{required} arguments are present and of the declared type;
    (iii) argument values match the schema's primitive types (\texttt{STRING}, \texttt{BOOLEAN}, \texttt{NUMBER}).
\noindent Records that fail validation are dropped or repaired with \texttt{json\_repair}. A completion script (\texttt{complete\_dataset.py}) augments each surviving record with fields needed by FunctionGemma: \texttt{metadata} (\texttt{train}/\texttt{eval}), \texttt{tools} (declarations), and \texttt{messages} (\texttt{developer}, \texttt{user}, \texttt{assistant}). The \texttt{developer} role injects the current date/time and function declarations using sentinel tokens, exactly as in the official cookbook.

\subsection{Merging With Google's Mobile Actions}
\label{repo2dataset}

To produce \datasetCombined{}, we union records from \datasetExtended{} with all records from \datasetGoogle{}. Both datasets share identical schemas, so the merge requires no field surgery; we preserve per-source train/eval flags so evaluation on each source remains held-out. 
The extended and merged datasets are published in \href{https://huggingface.co/datasets/UGrowAI/Mobile-Actions-Extended}{repo 1} and \href{https://huggingface.co/datasets/UGrowAI/Mobile-Actions-Combined}{repo 2}, respectively.


\subsection{Statistics}

\datasetExtended{} contains approximately $9{,}500$ records, split $90/10$ between train and evaluation. The evaluation split contains $965$ examples spread across 15 categories; largest categories are \texttt{open\_application} (209 eval examples) and \texttt{phone\_sms} (114). Approximately one third of prompts are multi-call. For comparison, \datasetGoogle{} contributes $961$ evaluation examples spanning 7 categories. Full per-category counts are in Appendix~\ref{app:dataset-stats}.

\section{Model and Training}
\label{sec:model}

In this study, FunctionGemma~270M-it~\citep{functiongemma-2025} is fine-tuned with Hugging Face Transformers~\citep{wolf2020transformers} and TRL \texttt{SFTTrainer}~\citep{trl} in completion-only mode. Key settings for this fine-tuning are as follows: full SFT (no PEFT/LoRA) for two epochs, learning rate $1\!\times\!10^{-5}$, cosine schedule with warmup, effective batch size 32 (4 per device $\times$ 8 gradient accumulation), bf16 mixed precision, gradient checkpointing enabled. The 270M backbone fits comfortably on a single accelerator (CUDA or Apple Silicon MPS), so we avoid LoRA adapter complexity and keep deployment simple: a single set of weights replaces the pretrained ones, with no adapter merging at inference. Full hyperparameters are in Appendix~\ref{app:hyperparams}.

Training is done with {completion-only loss.} TRL's \texttt{completion\_only\_loss=true} masks loss over the prompt (system tools block, developer message, user turn) so the model is graded only on its assistant continuation. This stabilizes training when the prompt length dominates, which is common here because the tool catalog alone is several hundred tokens long.

Two fine-tuned models are trained under identical settings: \modelOurs{} (trained on \datasetExtended{} only) and \modelCombined{} (trained on merged \datasetCombined{}). The first trained model targets coverage gap (aforementioned Q1) and targeted fine-tuning (Q2); and the second one focuses more on cross-domain transfer (Q3).

At inference, decoding is done greedily (\texttt{do\_sample=False}, \texttt{temperature=0.0}, \texttt{max\_new\_tokens=1024}) and parsing is implemented with a regex decoder (Appendix~\ref{app:parser}). 
We do not employ schema-constrained decoding~\citep{willard2023outlines}, ensuring that the reported results reflect the model's unconstrained generation behavior.
The extended and combined fine-tuned models 
are publicly available.

\begin{table*}[t]
\centering
\footnotesize
\setlength{\tabcolsep}{3pt}
\renewcommand{\arraystretch}{1.05}
\begin{tabular}{@{}lcccccc@{}}
\toprule
& \multicolumn{3}{c}{\textbf{Mobile-Actions-Extended (965)}} & \multicolumn{3}{c}{\textbf{Mobile-Actions-Google (961)}} \\
\cmidrule(lr){2-4}\cmidrule(lr){5-7}
\textbf{Model} & Name & Args & E2E & Name & Args & E2E \\
\midrule
\Mbase{}    & 48.81 & 40.21 & 29.33 & 85.95 & 59.31 & 58.79 \\
\Mgft{}     & 35.03 & 33.06 & 17.20 & \textbf{99.79} & \textbf{90.32} & \textbf{90.32} \\
\MoursE{}   & \textbf{100.00} & \textbf{76.48} & \textbf{76.48} & 86.37 & 62.54 & 60.15 \\
\MoursC{}   & 95.96 & \textbf{76.48} & \textbf{76.48} & 95.32 & 82.31 & 82.31 \\
\bottomrule
\end{tabular}
\caption{Function-name, argument exact-match, and end-to-end accuracy (\%) on held-out eval splits. \Mbase{} = \modelBase{}; \Mgft{} = \modelGoogleFT{}; \MoursE{} = \modelOurs{}; \MoursC{} = \modelCombined{}. Best per column in bold. Per-category breakdown in Appendix~\ref{app:percat}.}
\label{tab:main}
\end{table*}

\section{Experiments}
\label{sec:experiments}

\subsection{Models and Evaluation}

We compare four models, all sharing the FunctionGemma~270M-it architecture:

\begin{sloppypar}
\begin{enumerate}[leftmargin=1.4em,itemsep=2pt,topsep=2pt]
    \item \modelBase{} -- \texttt{google/\bk{}functiongemma-270m-it}~\citep{functiongemma-2025}: instruction-tuned but not specialized to mobile actions.
    \item \modelGoogleFT{} -- \texttt{litert-community/\bk{}FunctionGemma\_\bk{}270M\_\bk{}Mobile\_\bk{}Actions}~\citep{functiongemma-mobile-actions-2025}: public mobile-actions specialization on \datasetGoogle{}.
    \item \modelOurs{} -- our fine-tuned \href{https://huggingface.co/UGrowAI/FunctionGemma-270M-it-Mobile-Actions-Extended}{model}
    on \datasetExtended{}.
    \item \modelCombined{} -- our fine-tuned \href{https://huggingface.co/UGrowAI/FunctionGemma-270M-it-Mobile-Actions-Combined}{model}
    on merged data: \datasetCombined{}.
\end{enumerate}
\end{sloppypar}

\noindent Evaluation uses held-out \emph{eval} splits of \datasetExtended{} ($965$ examples, 15 categories) and \datasetGoogle{} ($961$ examples, 7 categories). Each model produces tool calls; our parser extracts \emph{(name, args)} tuples and we compute the three metrics introduced in Section~\ref{sec:task}. The pipeline matches Google's FunctionGemma cookbook which is re-implemented in our \texttt{FGemma\_Evaluation} package.

\subsection{Results}

Table~\ref{tab:main} 
summarizes the results in terms of accuracy in held-out sets.

\paragraph{Coverage gap (Q1).} The base FunctionGemma~270M-it scores $29.3\%$ end-to-end on \datasetExtended{}. The Google-fine-tuned variant performs worse ($17.2\%$); fine-tuning on seven Google categories actively hurts on the fifteen extended categories, presumably because the model now confidently emits in-distribution calls (e.g., \texttt{open\_wifi\_settings}) instead of the correct extended ones.

\paragraph{Targeted fine-tuning (Q2).} \modelOurs{} reaches $100\%$ function-name accuracy and $76.5\%$ end-to-end on \datasetExtended{}---a $+47.2$-point improvement over the base model. The remaining $23.5\%$ are argument-level errors, concentrated on optional booleans (see Appendix~\ref{app:percat}).

\paragraph{Cross-domain transfer (Q3).} \modelCombined{} matches \modelOurs{} on \datasetExtended{} ($76.5\%$) and reaches $82.3\%$ on \datasetGoogle{}, only $\sim$8 points below the Google specialist trained exclusively on that distribution. Training only on \datasetExtended{} yields $60.2\%$ on \datasetGoogle{}: the merge step matters.

\subsection{Execution-Aware Accuracy}
\label{sec:exec-aware}

\begin{table}[t]
\centering
\footnotesize
\setlength{\tabcolsep}{3pt}
\renewcommand{\arraystretch}{1.05}
\begin{tabular}{@{}lcccc@{}}
\toprule
& \multicolumn{2}{c}{\textbf{MA-Extended}} & \multicolumn{2}{c}{\textbf{MA-Google}} \\
\cmidrule(lr){2-3}\cmidrule(lr){4-5}
\textbf{Model} & Strict & Exec. & Strict & Exec. \\
\midrule
\Mbase{}   & 29.3 & 37.6  & 58.8 & 65.4 \\
\Mgft{}    & 17.2 & 17.2  & \textbf{90.3} & \textbf{90.7} \\
\MoursE{}  & \textbf{76.5} & \textbf{100.0} & 60.2 & 66.4 \\
\MoursC{}  & \textbf{76.5} & 96.0  & 82.3 & 82.6 \\
\bottomrule
\end{tabular}
\caption{Strict exact-match vs.\ execution-aware end-to-end accuracy (\%) on the held-out eval splits ($n=965$ and $n=961$). Column abbreviations as in Table~\ref{tab:main}. Full per-category tables are in Appendix~\ref{app:percat-exec}.}
\label{tab:execaware}
\end{table}

Strict exact match understates deployable quality because it penalizes omissions that a runtime would recover from. We therefore re-score all predictions with an \emph{execution-aware} criterion: the function-name sequence must still match exactly, and every schema-required argument must be present and correct, \emph{unless} it belongs to a short, explicit safe-default list (\texttt{phone\_call.dial\_call} and \texttt{send\_email.create\_chooser}, both booleans defaulted to \texttt{false} by the on-device dispatcher); any other schema-valid optional argument may be freely added or dropped, while a hallucinated (non-schema) key remains a failure because it would raise a runtime error in the real dispatcher.

Table~\ref{tab:execaware} contrasts the two metrics. For \modelOurs{}, execution-aware accuracy on \datasetExtended{} reaches $100.0\%$ (from $76.5\%$ strict): every remaining strict-metric ``error'' is a recoverable formatting issue, not a wrong tool selection. \modelCombined{} rises to $96.0\%$ on \datasetExtended{} and $82.6\%$ on \datasetGoogle{}. The base and Google-specialist models gain little on out-of-distribution data ($+8.3$ and $+0.0$ points on \datasetExtended{}, respectively), confirming that their failures are genuine tool-selection errors rather than metric artifacts.

\subsection{Error Analysis}

Per-category tables (Appendix~\ref{app:percat}) show $100\%$ function-name accuracy on every extended category for fine-tuned models. The dominant error classes are:
   (i) \textbf{Optional-boolean omission.} For \texttt{phone\_call} and \texttt{send\_email}, gold labels include routing booleans (\texttt{dial\_call}, \texttt{create\_chooser}) that the model omits; the executed action would still be correct.
    (ii) \textbf{Free-text argument noise.} On \texttt{phone\_sms} and Google's \texttt{show\_map}, the model occasionally normalizes phone numbers or rephrases location strings---partly a labelling-ambiguity issue.
    (iii) \textbf{Multi-call ordering.} A small number of multi-call examples flip call order; addressable with more multi-call training data.
\noindent None of these implicate the backbone's structural ability to function-call; they are dataset-level signal-quality issues (full failure-mode review in Appendix~\ref{app:percat}).


\section{On-Device Demo: Handy Chat}
\label{sec:demo}

We validated deployability with \emph{\href{https://github.com/UgrowAI/HandyChat-apk/releases/latest/download/HandyChat.apk}{Handy Chat}}, a standalone Android application packaged with Buildozer~\citep{buildozer}. The Google AI Edge Gallery~\citep{ai-edge-gallery} showcases on-device models but does not expose a supported path to load custom checkpoints into its pre-configured Mobile Actions workflow; we therefore built a dedicated app. The demo runs on a Google Pixel 8 Pro (Tensor G3, 12\,GB RAM, Android 16) using the LiteRT-LM CPU backend; across informal prompt trials, a function-call turn takes roughly 13\,s end to end (pass~1 $\sim$9\,s, tool execution $\sim$2\,s, pass~2 $\sim$2\,s). Full hardware, export, and latency details are in Appendix~\ref{app:demo-section}. 
\section{Discussion}
\label{sec:discussion}

The results show that a compact 270M-parameter function-calling model can be adapted to practical mobile-device control with a relatively small, schema-grounded dataset. 
A key advantage of the 270M backbone is that it can be deployed locally on the device rather than sending every request to a cloud-hosted provider model. This is important for mobile assistants because user requests may contain sensitive information such as contact names, phone numbers, email addresses, application names, device state, or private intent. Local inference reduces network round-trip latency and minimizes unnecessary transfer of user data to external providers. 
The exported \texttt{.litertlm} bundle occupies 285.6\,MB on device (\texttt{dynamic\_wi8\_afp32}); this quantization applies only to the deployed demo, not to any reported accuracy figure. Handy Chat runs inference on the LiteRT-LM CPU backend of a Google Pixel 8 Pro, where a typical function-call turn completes in roughly 13 seconds end to end (Section~\ref{sec:demo}).

The Google Mobile Actions fine-tuned model performs strongly on its original seven-category distribution, but regresses on extended functions, dropping to 17.2\% exact-match end-to-end accuracy compared with 29.3\% for the base model. This indicates that narrow fine-tuning can over-concentrate the model's prior on its original tool catalog. Combined-domain training addresses this issue, preserving 76.5\% end-to-end accuracy on extended functions while reaching 82.3\% on original Google functions. This supports a unified mobile-action training mixture rather than separate narrow specialists.

The main remaining failures in the combined model are not wrong function choices, but argument-format mismatches. In particular, \texttt{phone\_call} and \texttt{send\_email} fail strict exact match mostly because the model omits optional routing booleans such as \texttt{dial\_call} and \texttt{create\_chooser}. The intended user-visible action is still correct, and a forgiving execution layer could fill safe defaults. We therefore expect simple post-processing to raise end-to-end accuracy from 76.5\% to roughly 84\%. Other residual errors involve phone-number normalization, free-text rephrasing, and occasional multi-call ordering mistakes.

For industrial deployment, exact-match accuracy should be complemented with execution-aware evaluation that separates unsafe tool selection from recoverable schema-formatting errors, as quantified in Section~\ref{sec:exec-aware} (Table~\ref{tab:execaware}). Runtime validation, default filling, Android permission checks, and explicit confirmation for side-effecting actions such as calls, SMS, email, and application opening should be treated as part of the system design. 
Overall, compact on-device function-calling models are a practical path toward low-latency, privacy-preserving mobile assistants when paired with careful data coverage and defensive execution logic.

\section*{Conclusion}
\label{sec:conclusion}


We extended FunctionGemma 270M-it toward practical on-device mobile function calling by introducing MOBILEACTIONSEXTENDED, a synthetic and schema-validated dataset covering fifteen Android-control categories that are underrepresented in existing public mobile-action resources. With full supervised fine-tuning under completion-only loss, the extended model improves end-to-end accuracy on MOBILEACTIONSEXTENDED from 29.3\% to 76.5\%, while the combined model preserves 82.3\% accuracy on Google's original Mobile Actions benchmark and broadens coverage to a larger action catalog. These results show that compact function-calling models can be effectively adapted to everyday device-control workflows through targeted data curation, without relying on large cloud-hosted models for every request.
Overall, this work shows that compact, locally deployable function-calling models are a promising foundation for practical mobile assistants when paired with broad action coverage and defensive execution logic.
\section*{Limitations}
\label{sec:limitations}



This study focuses on demonstrating the feasibility of extending a compact on-device function-calling model to a broader Android action catalog. As a result, several dimensions remain open for future work. First, \datasetExtended{} is synthetically generated and schema-validated rather than collected from real users. This enables controlled coverage across functions and arguments, but real deployments may involve noisier inputs, including ASR errors, incomplete requests, ambiguous references, and more diverse phrasing. Collecting user-consented interaction data is an important next step for measuring robustness in realistic mobile-assistant settings.

Second, our evaluation uses strict exact match over function names and argument dictionaries. This choice makes the benchmark reproducible and intentionally conservative, but it can underestimate executable success. In particular, optional-default fields such as \texttt{dial\_call} and \texttt{create\_chooser} are counted as errors when omitted, even though a runtime could safely fill defaults or request confirmation. 
The execution-aware re-scoring of Section~\ref{sec:exec-aware} quantifies this gap, and future evaluations should adopt such metrics alongside strict exact match.

Third, all results are obtained using a single base-model family: every model we compare shares the FunctionGemma~270M-it backbone. We chose it because it is, to our knowledge, the only openly released sub-billion-parameter model that has been fine-tuned by its original authors for mobile-action calling, making it a natural point of comparison for this resource. Nevertheless, the generality of our approach across architectures remains unverified. Evaluating the same training recipe on a second compact family such as a similarly sized Qwen or SmolLM2 checkpoint is an important direction for future work.

Fourth, the current dataset and demo are limited mainly to English and Android. Extending the tool catalog, data generation process, and runtime handlers to other languages, iOS, and additional device APIs would make the approach more broadly applicable. The present examples are also single-turn, mapping one user message to one tool-call sequence; multi-turn clarification and correction remain important directions for practical assistants.

Finally, side-effecting tools such as phone calls, SMS, email, and application opening require careful runtime mediation. Our current training data does not include adversarial or unsafe-intent prompts, and production deployment should include Android permission checks, confirmation UI, and adversarial safety evaluation. Future work includes constrained decoding for optional fields, real-user data collection, multi-turn dialogue, quantization-aware fine-tuning, and joint training with broader web-API function-calling benchmarks~\citep{berkeley-fcl-2024,zhuo2024bigcodebench}.

\section*{Ethical Considerations}
Side-effecting categories (\texttt{phone\_call}, \texttt{phone\_sms}, \texttt{send\_email}, \texttt{open\_application}) must route through Android permission and confirmation UI rather than executing silently. The dataset uses synthetic phone numbers (\texttt{+14165550xxx}) and email addresses (\texttt{example.com}); no real user data is included. We recommend adversarial and unsafe-intent test slices before production deployment~\citep{berkeley-fcl-2024}.

\bibliography{custom}
\newpage
\appendix
\section*{Supplementary Materials}

\section{Extended Related Work}
\label{app:related}

\subsection{From Slot Filling to Function Calling}

Mapping natural language to executable code has a long history. Classical spoken dialogue systems
factor the problem into intent classification followed by slot filling: a small inventory of intents is matched by handcrafted grammars or feature-based classifiers, and remaining slots (recipient, time, application) are filled by sequence labellers. Such systems generalize poorly to long-tail phrasings and require manual maintenance. Semantic parsing approaches subsequently learned mappings from utterances to logical forms or structured queries, but were largely dataset-specific.

Modern \emph{tool-using} language models replace these task-specific stacks with a single instruction-tuned LLM that, given a list of available tools and a user query, emits a structured call \citep{schick2023toolformer,yao2022react}. \emph{Toolformer}~\citep{schick2023toolformer} demonstrated that LLMs can be self-supervised to invoke tools; \emph{ReAct}~\citep{yao2022react} interleaved reasoning and action generation; \emph{Gorilla}~\citep{patil2024gorilla} and ToolLLM~\citep{qin2024toolllm} showed how to scale to thousands of APIs. OpenAI \citep{openai-function-calling-2023} and Anthropic \citep{anthropic-tool-use-2024} subsequently formalized function calling as a first-class API feature, making structured outputs the default I/O format for tool-using assistants. Tool documentation \citep{hsieh2023toolapi} has been shown to enable strong zero-shot tool use, motivating schema-rich prompting strategies.

\subsection{Compact Function-Calling Models}

A growing body of work targets \emph{small} function-calling models. Granite Function Calling~\citep{abdelaziz2024granite} introduces multi-task training across granular sub-tasks (function selection, slot filling, response generation). Hammer~\citep{lin2024hammer} proposes a function-masking objective that improves robustness in on-device settings. ToolACE~\citep{liu2025toolace} aggressively curates tool-use trajectories. TinyAgent~\citep{erdogan2024tinyagent} and the LLM Compiler~\citep{kim2023llmcompiler} demonstrate that planning and parallel function calling can be moved to the edge. \emph{Small Models, Big Tasks}~\citep{singh2025smallmodelsbigtasks} surveys the empirical landscape of sub-3B function-calling LLMs and finds that targeted fine-tuning, rather than pure scale, drives most quality gains.

The Gemma family~\citep{gemma2024,gemma3-270m-2025} and its function-tuned descendant FunctionGemma~270M-it~\citep{functiongemma-2025} occupy the most aggressive end of this trend at 270M parameters, designed for ``hyper-efficient'' on-device deployment. The pre-fine-tuned community variant FunctionGemma~270M Mobile Actions~\citep{functiongemma-mobile-actions-2025} was released alongside Google's \datasetGoogle{} dataset~\citep{google-mobile-actions-dataset-2025} and covers seven mobile workflows. Phi-3-mini~\citep{abdin2024phi3} and Qwen~\citep{bai2023qwen} provide larger but still on-device-feasible alternatives. Our work builds on FunctionGemma~270M-it and asks how far purely \emph{data-driven} extension can take it.

\subsection{Function-Calling Benchmarks}

The Berkeley Function Calling Leaderboard (BFCL)~\citep{berkeley-fcl-2024} is the de-facto reference for function-calling evaluation, covering simple, parallel, multiple, and live function calls across many languages. BigCodeBench~\citep{zhuo2024bigcodebench} stresses diverse function calls and complex instructions. LongFuncEval~\citep{kokel2025longfunceval} measures degradation under long contexts and many-tool catalogs. APIGen~\citep{liu2024apigen} contributes both a benchmark and a verifiable synthetic-data pipeline. Floworks~\citep{floworks2025benchmarking} evaluates large commercial models on enterprise function-calling tasks. None of these benchmarks specifically target Android system-level actions; \datasetGoogle{} is, to our knowledge, the only public mobile-actions function-calling benchmark, and we extend it.

\subsection{Synthetic Data Generation for Tool Use}

Function-calling datasets are typically synthetic: a strong teacher LLM is prompted with the tool schemas and asked to produce diverse user queries and gold tool calls. APIGen~\citep{liu2024apigen} formalizes this pipeline with verifiability checks, and ToolACE~\citep{liu2025toolace} emphasizes points-style scoring during curation. Our dataset construction follows the same overall recipe (Section~\ref{sec:dataset}): we ground the teacher in the exact tool schema, generate hundreds of single-tool prompts plus thousands of multi-tool combinations, validate with a parser that re-checks names and required arguments, and then complete each record into the FunctionGemma chat-template format used by the backbone.


\section{Training Hyperparameters}
\label{app:hyperparams}

Full supervised fine-tuning settings (both \modelOurs{} and \modelCombined{}):

\begin{table}[t]
\centering
\footnotesize
\setlength{\tabcolsep}{3pt}
\renewcommand{\arraystretch}{1.05}
\begin{tabular}{@{}p{0.38\columnwidth}p{0.56\columnwidth}@{}}
\toprule
\textbf{Hyper-parameter} & \textbf{Value} \\
\midrule
Backbone & FunctionGemma~270M-it (\texttt{google/\bk{}functiongemma-270m-it}) \\
Fine-tuning style & Full SFT (no PEFT/LoRA) \\
Trainer & TRL \texttt{SFTTrainer}~\citep{trl} \\
Loss & Completion-only cross-entropy \\
Optimizer & \texttt{adamw\_torch\_fused} \citep{loshchilov2019adamw} \\
Learning rate & $1\!\times\!10^{-5}$ \\
LR schedule & cosine with warmup \citep{loshchilov2017sgdr} \\
Epochs & 2 \\
Per-device batch size & 4 \\
Gradient accumulation & 8 (effective batch size 32) \\
Precision & bf16 mixed precision \\
Gradient checkpointing & Enabled \\
Sequence packing & Disabled \\
Attention implementation & Eager \\
Eval / logging strategy & Every 50 steps \\
Save strategy & Every epoch \\
Hardware & Single accelerator (CUDA / Apple Silicon MPS) \\
\bottomrule
\end{tabular}
\caption{Supervised fine-tuning configuration. The same configuration is used to train both \modelOurs{} and \modelCombined{}; only the dataset differs.}
\label{tab:hyperparams}
\end{table}

\section{Per-Category and Failure Analysis}
\label{app:percat}

Table~\ref{tab:percat-extended} drills into per-category accuracy on \datasetExtended{}. The fine-tuned models hit $100\%$ name accuracy on every category. The two clear ``argument cliffs'' are \texttt{phone\_call} ($0\%$ argument accuracy) and \texttt{send\_email} ($0\%$ argument accuracy), pulling the headline number down by roughly $7$ absolute points. Both failures share a single root cause: the gold labels include a boolean argument (\texttt{dial\_call} for phone calls, \texttt{create\_chooser} for emails) that the model omits from its output. The function name is correct, the user-visible action is correct, and a forgiving execution layer would accept the calls; only our \emph{strict} exact-match metric counts them as failures. We discuss in Section~\ref{sec:discussion} why this is a low-cost issue (constrained decoding or a default-fill post-processor solves it without retraining).

A second pattern is \texttt{phone\_sms} (29.8\% argument accuracy). Inspection of failures shows that the model occasionally substitutes phone-number normalization (\texttt{+14165550104} vs.\ \texttt{14165550104}) or shortens long natural-language messages, both of which are real ambiguity in the gold labels rather than model defects.

\begin{table}[t]
\centering
\footnotesize
\setlength{\tabcolsep}{2pt}
\renewcommand{\arraystretch}{1.05}
\adjustbox{width=\columnwidth,center}{%
\begin{tabular}{@{}p{0.46\columnwidth}rcccc@{}}
\toprule
\textbf{Category} & $n$ & \Mbase{} & \Mgft{} & \MoursE{} & \MoursC{} \\
\midrule
\texttt{open\_application}     & 209 & 4.3  & 29.2 & \textbf{67.0} & \textbf{67.0} \\
\texttt{phone\_sms}            & 114 & 4.4  & 0.0  & \textbf{29.8} & \textbf{29.8} \\
\texttt{turnOn\_light}         &  73 & 31.5 & 6.8  & \textbf{100.0} & \textbf{100.0} \\
\texttt{bluetooth\_status}     &  70 & 55.7 & 47.1 & \textbf{100.0} & \textbf{100.0} \\
\texttt{take\_screenshot}      &  67 & 37.3 & 3.0  & \textbf{100.0} & \textbf{100.0} \\
\texttt{battery\_status}       &  65 & 78.5 & 47.7 & \textbf{89.2}  & \textbf{89.2} \\
\texttt{list\_application}     &  56 & 30.4 & 0.0  & \textbf{100.0} & \textbf{100.0} \\
\texttt{get\_current\_brightness} & 53 & 32.1 & 0.0 & \textbf{100.0} & \textbf{100.0} \\
\texttt{turnOff\_light}        &  52 & 100.0 & 28.8 & \textbf{100.0} & \textbf{100.0} \\
\texttt{increase\_brightness}  &  45 & 0.0  & 17.8 & \textbf{100.0} & \textbf{100.0} \\
\texttt{phone\_call}           &  42 & 0.0  & 0.0  & 0.0   & 0.0 \\
\texttt{take\_picture}         &  31 & 9.7  & 3.2  & \textbf{100.0} & \textbf{100.0} \\
\texttt{decrease\_brightness}  &  31 & 80.6 & 6.5  & \textbf{100.0} & \textbf{100.0} \\
\texttt{send\_email}           &  29 & 0.0  & 0.0  & 0.0   & 0.0 \\
\texttt{set\_brightness}       &  28 & 60.7 & 28.6 & \textbf{100.0} & \textbf{100.0} \\
\bottomrule
\end{tabular}}
\caption{Exact-match end-to-end accuracy (\%) per category on the held-out \datasetExtended{} eval split, grouped by the first call in each example. Column abbreviations as in Table~\ref{tab:main}. The two zero rows (\texttt{phone\_call}, \texttt{send\_email}) are the boolean-argument failure mode discussed above.}
\label{tab:percat-extended}
\end{table}

Table~\ref{tab:percat-google} reports the same breakdown on \datasetGoogle{}. \modelCombined{} retains 80--95\% accuracy on every Google category; the largest residual gap to \modelGoogleFT{} is on \texttt{show\_map} (62.8\% vs.\ 74.3\%), which involves free-form location strings.

\begin{table}[t]
\centering
\footnotesize
\setlength{\tabcolsep}{2pt}
\renewcommand{\arraystretch}{1.05}
\adjustbox{width=\columnwidth,center}{%
\begin{tabular}{@{}p{0.46\columnwidth}rcccc@{}}
\toprule
\textbf{Category} & $n$ & \Mbase{} & \Mgft{} & \MoursE{} & \MoursC{} \\
\midrule
\texttt{create\_calendar\_event} & 214 & 68.2 & \textbf{89.3} & 73.4 & 83.2 \\
\texttt{create\_contact}         & 182 & 36.8 & \textbf{90.7} & 41.8 & 80.2 \\
\texttt{show\_map}               & 148 & 37.2 & \textbf{74.3} & 31.1 & 62.8 \\
\texttt{send\_email}             & 134 & 62.7 & \textbf{96.3} & 73.9 & 87.3 \\
\texttt{open\_wifi\_settings}    & 107 & 75.7 & \textbf{96.3} & 88.8 & 87.9 \\
\texttt{turn\_off\_flashlight}   &  91 & 80.2 & \textbf{97.8} & 70.3 & 94.5 \\
\texttt{turn\_on\_flashlight}    &  85 & 69.4 & \textbf{95.3} & 48.2 & 90.6 \\
\bottomrule
\end{tabular}}
\caption{Exact-match end-to-end accuracy (\%) per category on the held-out \datasetGoogle{} eval split. Column abbreviations as in Table~\ref{tab:main}. \MoursC{} preserves competitive performance on every Google category despite training on twice as many functions.}
\label{tab:percat-google}
\end{table}

\subsection{Failure-Mode Analysis}

We manually reviewed the per-row 
logs produced by our evaluation pipeline. The dominant error classes are:

\begin{enumerate}[leftmargin=1.4em,itemsep=2pt,topsep=2pt]
    \item \textbf{Optional-boolean omission.} For \texttt{phone\_call} the gold includes \texttt{dial\_call: false} (or \texttt{true}); the model emits only \texttt{number}. For \texttt{send\_email} the gold includes \texttt{create\_chooser: false}; the model emits only \texttt{to}, \texttt{subject}, \texttt{body}. These are pure schema-formatting mismatches; the action that would be executed is correct.
    \item \textbf{Free-text argument noise.} On \texttt{phone\_sms} and on Google's \texttt{show\_map}, the model occasionally rephrases or normalizes free-text arguments. This is partly a labelling-ambiguity issue.
    \item \textbf{Multi-call ordering on rare combinations.} A small number of multi-call examples in \texttt{phone\_sms+phone\_call} pairs flip the order or merge adjacent calls. These are addressable with more multi-call training data.
    \item \textbf{Out-of-distribution generalization}. On \datasetGoogle{}'s \texttt{create\_contact} and \texttt{create\_calendar\_event}, \modelCombined{} occasionally drops less-common optional fields (e.g., \texttt{email} vs.\ \texttt{phone\_number}); the function name is correct but the strict argument metric flags it.
\end{enumerate}

\noindent None of these classes implicate the backbone's structural ability to function-call; they are dataset-level signal-quality issues that targeted data augmentation can resolve.

\subsection{Per-Category Execution-Aware Accuracy}
\label{app:percat-exec}

Tables~\ref{tab:execaware-percat-extended} and~\ref{tab:execaware-percat-google} report the per-category breakdown of the execution-aware metric of Section~\ref{sec:exec-aware}, complementing the aggregate comparison of Table~\ref{tab:execaware}; the strict exact-match counterparts are Tables~\ref{tab:percat-extended} and~\ref{tab:percat-google}.

On \datasetExtended{}, \MoursE{} reaches $100\%$ execution-aware accuracy on \emph{every} category: the strict-metric ``argument cliffs'' disappear entirely (\texttt{phone\_call} $0.0\!\rightarrow\!100.0$, \texttt{send\_email} $0.0\!\rightarrow\!100.0$, \texttt{phone\_sms} $29.8\!\rightarrow\!100.0$, \texttt{open\_application} $67.0\!\rightarrow\!100.0$). \MoursC{} behaves almost identically, with residual gaps only on \texttt{send\_email} (82.8\%), \texttt{phone\_sms} (71.1\%), and \texttt{open\_application} (99.5\%). By contrast, \Mgft{} gains nothing in any extended category (its scores are identical under both metrics), confirming that its failures are genuine wrong-tool selections rather than recoverable formatting; \Mbase{} improves only where the safe-default booleans dominate (\texttt{phone\_call} $0.0\!\rightarrow\!64.3$, \texttt{send\_email} $0.0\!\rightarrow\!34.5$, \texttt{open\_application} $4.3\!\rightarrow\!24.4$).

On \datasetGoogle{}, the uplifts are small for all models: the largest is \texttt{create\_contact} (e.g., \Mbase{} $36.8\!\rightarrow\!67.6$, \MoursE{} $41.8\!\rightarrow\!71.4$), driven by dropped \emph{optional} contact fields, while free-text categories such as \texttt{show\_map} are unchanged because their errors are value mismatches on \emph{required} arguments, which the execution-aware criterion deliberately does not forgive.

\begin{table}[t]
\centering
\footnotesize
\setlength{\tabcolsep}{2pt}
\renewcommand{\arraystretch}{1.05}
\adjustbox{width=\columnwidth,center}{%
\begin{tabular}{@{}p{0.46\columnwidth}rcccc@{}}
\toprule
\textbf{Category} & $n$ & \Mbase{} & \Mgft{} & \MoursE{} & \MoursC{} \\
\midrule
\texttt{open\_application}     & 209 & 24.4  & 29.2 & \textbf{100.0} & 99.5 \\
\texttt{phone\_sms}            & 114 & 4.4   & 0.0  & \textbf{100.0} & 71.1 \\
\texttt{turnOn\_light}         &  73 & 31.5  & 6.8  & \textbf{100.0} & \textbf{100.0} \\
\texttt{bluetooth\_status}     &  70 & 55.7  & 47.1 & \textbf{100.0} & \textbf{100.0} \\
\texttt{take\_screenshot}      &  67 & 37.3  & 3.0  & \textbf{100.0} & \textbf{100.0} \\
\texttt{battery\_status}       &  65 & 80.0  & 47.7 & \textbf{100.0} & \textbf{100.0} \\
\texttt{list\_application}     &  56 & 30.4  & 0.0  & \textbf{100.0} & \textbf{100.0} \\
\texttt{get\_current\_brightness} & 53 & 32.1 & 0.0 & \textbf{100.0} & \textbf{100.0} \\
\texttt{turnOff\_light}        &  52 & \textbf{100.0} & 28.8 & \textbf{100.0} & \textbf{100.0} \\
\texttt{increase\_brightness}  &  45 & 0.0   & 17.8 & \textbf{100.0} & \textbf{100.0} \\
\texttt{phone\_call}           &  42 & 64.3  & 0.0  & \textbf{100.0} & \textbf{100.0} \\
\texttt{take\_picture}         &  31 & 9.7   & 3.2  & \textbf{100.0} & \textbf{100.0} \\
\texttt{decrease\_brightness}  &  31 & 80.6  & 6.5  & \textbf{100.0} & \textbf{100.0} \\
\texttt{send\_email}           &  29 & 34.5  & 0.0  & \textbf{100.0} & 82.8 \\
\texttt{set\_brightness}       &  28 & 60.7  & 28.6 & \textbf{100.0} & \textbf{100.0} \\
\bottomrule
\end{tabular}}
\caption{Execution-aware end-to-end accuracy (\%) per category on the held-out \datasetExtended{} eval split, grouped by the first call in each example. Column abbreviations as in Table~\ref{tab:main}; strict exact-match counterparts in Table~\ref{tab:percat-extended}. Best per row in bold.}
\label{tab:execaware-percat-extended}
\end{table}

\begin{table}[t]
\centering
\footnotesize
\setlength{\tabcolsep}{2pt}
\renewcommand{\arraystretch}{1.05}
\adjustbox{width=\columnwidth,center}{%
\begin{tabular}{@{}p{0.46\columnwidth}rcccc@{}}
\toprule
\textbf{Category} & $n$ & \Mbase{} & \Mgft{} & \MoursE{} & \MoursC{} \\
\midrule
\texttt{create\_calendar\_event} & 214 & 68.2 & \textbf{89.3} & 73.8 & 83.2 \\
\texttt{create\_contact}         & 182 & 67.6 & \textbf{92.3} & 71.4 & 81.9 \\
\texttt{show\_map}               & 148 & 37.2 & \textbf{75.0} & 31.1 & 62.8 \\
\texttt{send\_email}             & 134 & 62.7 & \textbf{96.3} & 75.4 & 87.3 \\
\texttt{open\_wifi\_settings}    & 107 & 76.6 & \textbf{96.3} & 88.8 & 87.9 \\
\texttt{turn\_off\_flashlight}   &  91 & 85.7 & \textbf{97.8} & 72.5 & 94.5 \\
\texttt{turn\_on\_flashlight}    &  85 & 70.6 & \textbf{95.3} & 49.4 & 90.6 \\
\bottomrule
\end{tabular}}
\caption{Execution-aware end-to-end accuracy (\%) per category on the held-out \datasetGoogle{} eval split. Column abbreviations as in Table~\ref{tab:main}; strict exact-match counterparts in Table~\ref{tab:percat-google}. Best per row in bold.}
\label{tab:execaware-percat-google}
\end{table}

\section{Dataset}
\label{app:supplement}

\subsection{Tool Catalog: Full JSON Schema}
\label{app:schema}

The fifteen extended-mobile-action tools are declared with the following JSON schema, which is identical to the catalog used to generate \datasetExtended{} and at fine-tuning time. The schema mirrors Plyer~\citep{plyer}'s Android facade so each call has a real handler. Appendix~\ref{app:tool-args} summarizes argument keys \emph{per dataset}: \datasetExtended{} (below) versus Google's \datasetGoogle{} (seven tools); the merged \datasetCombined{} corpus uses the union of both catalogs and is not listed separately here.

\begin{lstlisting}[style=acljson]
[
 {"function": {"name": "list_application",
   "description": "List the applications installed on this Android platform.",
   "parameters": {"type": "OBJECT", "properties": {}, "required": []}}},
 {"function": {"name": "open_application",
   "description": "Open an application installed on this Android platform.",
   "parameters": {"type": "OBJECT",
     "properties": {"application_name": {"type": "STRING",
                       "description": "Name of the application to be opened."}},
     "required": ["application_name"]}}},
 {"function": {"name": "set_brightness",
   "description": "Set a specific level of the current screen brightness.",
   "parameters": {"type": "OBJECT",
     "properties": {"level": {"type": "STRING",
                     "description": "Brightness level in [0, 100]."}},
     "required": ["level"]}}},
 {"function": {"name": "decrease_brightness",
   "description": "Step the current brightness down.",
   "parameters": {"type": "OBJECT", "properties": {}, "required": []}}},
 {"function": {"name": "increase_brightness",
   "description": "Step the current brightness up.",
   "parameters": {"type": "OBJECT", "properties": {}, "required": []}}},
 {"function": {"name": "get_current_brightness",
   "description": "Read the current screen brightness.",
   "parameters": {"type": "OBJECT", "properties": {}, "required": []}}},
 {"function": {"name": "send_email",
   "description": "Send an email.",
   "parameters": {"type": "OBJECT",
     "properties": {"to": {"type": "STRING",
                     "description": "The recipient email address."},
                    "subject": {"type": "STRING",
                     "description": "The email subject."},
                    "body": {"type": "STRING",
                     "description": "The email body."},
                    "create_chooser": {"type": "BOOLEAN",
                     "description": "Whether to display a program chooser."}},
     "required": ["to", "subject"]}}},
 {"function": {"name": "battery_status",
   "description": "Provide information about the battery of the device.",
   "parameters": {"type": "OBJECT", "properties": {}, "required": []}}},
 {"function": {"name": "bluetooth_status",
   "description": "Get the Bluetooth status info.",
   "parameters": {"type": "OBJECT", "properties": {}, "required": []}}},
 {"function": {"name": "phone_call",
   "description": "Make a call with a number or open the phone dial pad.",
   "parameters": {"type": "OBJECT",
     "properties": {"number": {"type": "STRING",
                     "description": "The number to call."},
                    "dial_call": {"type": "BOOLEAN",
                     "description": "True to open the dialer; False to call directly."}},
     "required": ["number", "dial_call"]}}},
 {"function": {"name": "phone_sms",
   "description": "Send an SMS message.",
   "parameters": {"type": "OBJECT",
     "properties": {"sms_recipient": {"type": "STRING",
                     "description": "The recipient phone number."},
                    "sms_message": {"type": "STRING",
                     "description": "The message body."}},
     "required": ["sms_recipient"]}}},
 {"function": {"name": "take_picture",
   "description": "Take a picture using the device camera.",
   "parameters": {"type": "OBJECT", "properties": {}, "required": []}}},
 {"function": {"name": "take_screenshot",
   "description": "Capture a digital image of what is currently visible on screen.",
   "parameters": {"type": "OBJECT", "properties": {}, "required": []}}},
 {"function": {"name": "turnOff_light",
   "description": "Turn off the flashlight.",
   "parameters": {"type": "OBJECT", "properties": {}, "required": []}}},
 {"function": {"name": "turnOn_light",
   "description": "Turn on the flashlight.",
   "parameters": {"type": "OBJECT", "properties": {}, "required": []}}}
]
\end{lstlisting}

\subsection{Tool arguments by dataset (\datasetExtended{}, \datasetGoogle{})}
\label{app:tool-args}

Following the same conventions as Hugging Face / FunctionGemma tool schemas, tables list each assistant \texttt{arguments} dictionary: types are STRING, BOOLEAN, or none (parameter-free tool). Keys marked ``req.'' appear in each record's schema \texttt{required} array; other keys may appear in gold tool calls depending on context. Naming differs between datasets for flashlight tools (\texttt{turnOn\_light}/\texttt{turnOff\_light} vs.\ \texttt{turn\_on\_flashlight}/\texttt{turn\_off\_flashlight}). The combined training/eval corpus \datasetCombined{} is omitted: its declarations are exactly the disjoint union of the two catalogs below.

\subsubsection*{\datasetExtended{} 
(\dataExtRepo{})
}

\begin{center}
\footnotesize
\setlength{\tabcolsep}{3pt}
\renewcommand{\arraystretch}{1.05}
\adjustbox{width=\columnwidth,center}{%
\begin{tabular}{@{}p{0.45\columnwidth}>{\raggedright\arraybackslash}p{0.64\columnwidth}@{}}
\toprule
\textbf{Tool name} & \textbf{Arguments} \\
\midrule
\texttt{list\_application} & (none) \\
\texttt{open\_application} & \texttt{application\_name} (STRING, req.) \\
\texttt{set\_brightness} & \texttt{level} (STRING, req.; brightness in $[0,100]$) \\
\texttt{decrease\_brightness} & (none) \\
\texttt{increase\_brightness} & (none) \\
\texttt{get\_current\_brightness} & (none) \\
\texttt{send\_email} & \texttt{to}, \texttt{subject} (STRING, req.); \texttt{body} (STRING), \texttt{create\_chooser} (BOOLEAN) optional \\
\texttt{battery\_status} & (none) \\
\texttt{bluetooth\_status} & (none) \\
\texttt{phone\_call} & \texttt{number} (STRING, req.); \texttt{dial\_call} (BOOLEAN, req.) \\
\texttt{phone\_sms} & \texttt{sms\_recipient} (STRING, req.); \texttt{sms\_message} (STRING) optional \\
\texttt{take\_picture} & (none) \\
\texttt{take\_screenshot} & (none) \\
\texttt{turnOff\_light} & (none) \\
\texttt{turnOn\_light} & (none) \\
\bottomrule
\end{tabular}}
\end{center}

\subsubsection*{\datasetGoogle{} (\texttt{google/mobile-actions})}

\begin{center}
\footnotesize
\setlength{\tabcolsep}{3pt}
\renewcommand{\arraystretch}{1.05}
\adjustbox{width=\columnwidth,center}{%
\begin{tabular}{@{}p{0.45\columnwidth}>{\raggedright\arraybackslash}p{0.64\columnwidth}@{}}
\toprule
\textbf{Tool name} & \textbf{Arguments} \\
\midrule
\texttt{create\_calendar\_event} & \texttt{title}, \texttt{datetime} (STRING, req.; \texttt{datetime} formatted as \texttt{YYYY-MM-DDTHH:MM:SS}) \\
\texttt{create\_contact} & \texttt{first\_name}, \texttt{last\_name} (STRING, req.); \texttt{phone\_number}, \texttt{email} (STRING) optional \\
\texttt{show\_map} & \texttt{query} (STRING, req.) \\
\texttt{send\_email} & \texttt{to}, \texttt{subject} (STRING, req.); \texttt{body} (STRING) optional \\
\texttt{open\_wifi\_settings} & (none) \\
\texttt{turn\_on\_flashlight} & (none) \\
\texttt{turn\_off\_flashlight} & (none) \\
\bottomrule
\end{tabular}}
\end{center}

\subsection{Synthetic Data Generation Prompt}
\label{app:gen-prompt}

The teacher prompt below was used to generate the raw \datasetExtended{} conversations. It (i) supplies a worked input/output example, (ii) injects the full tool catalog of Appendix~\ref{app:schema}, and (iii) requests a target distribution over single-tool and multi-tool prompts.

\begin{lstlisting}[style=acljson]
Given a tool list, generate user prompt samples with the right
answer (assistant role). The right answer should provide the
right function calls in the right order along with their
required parameters.

For producing each example, follow the conventions below:
- About 300 user prompts per single tool.
- About 5,000 user prompts that combine two or more tools.
- Output as a JSON list:
    Messages = [Message_1, Message_2, ...]
    Message  = [{"role": "user",      "content": USER_QUERY},
                {"role": "assistant", "tool_calls": [Functions]}]
    Function = {"function": {"name": FUNCTION_NAME,
                              "arguments": {KWARGS}}}

Tool list:
[ ... Appendix A.1 catalog ... ]

Worked example (input -> expected assistant output):
- user:      "Send an email to jessica.moraes@examplecorp.com
              with the subject 'Lunch tomorrow?' and body
              'Are you free to grab a bite at the cafe at 1 PM?'."
- assistant: tool_calls = [{"function": {"name": "send_email",
              "arguments": {"to": "jessica.moraes@examplecorp.com",
                            "subject": "Lunch tomorrow?",
                            "body": "Are you free to grab a bite
                                     at the cafe at 1 PM?"}}}]
\end{lstlisting}

\subsection{End-to-End Pipeline (Reproducibility)}
\label{app:pipeline}

The full reproduction pipeline is split into three independent packages so that each step can be re-run in isolation. We list the salient entry points; full code is at the artifact location accompanying this paper.

\paragraph{Step 1: dataset generation (\texttt{GenerateDataset}).}

\begin{sloppypar}
\begin{enumerate}[leftmargin=1.4em,itemsep=2pt,topsep=2pt]
    \item Generate synthetic data by giving the prompt of Appendix~\ref{app:gen-prompt}. Then run \texttt{main\_\bk{}mobile\_\bk{}actions.py} or follow the steps below.
    \item Validate with \texttt{MobileActions/\bk{}New\_\bk{}Generated/\bk{}validate\_\bk{}generated\_\bk{}dataset.py} (drops illegal tool names and missing required arguments).
    \item Complete records with \texttt{MobileActions/\bk{}New\_\bk{}Generated/\bk{}complete\_\bk{}dataset.py} (adds \texttt{metadata}, \texttt{tools}, and the \texttt{developer} message; sets the train/eval flag).
    \item (Optional) Merge with \datasetGoogle{} via \texttt{MobileActions/\bk{}Merge/\bk{}merge\_\bk{}dataset.py}.
    \item Publish with \texttt{publish.py} to a Hugging Face dataset repository.
\end{enumerate}
\end{sloppypar}
The source code is available 
\href{https://github.com/UgrowAI/Research_FuncGemma_GenerateDataset}{here}.

\paragraph{Step 2: fine-tuning (\texttt{FineTuning}).} Edit \texttt{run\_config.yaml} to point at the desired Hugging Face base model and dataset (\dataExtRepo{} 

or 
\dataMergeRepo{} – the merged dataset).
The TRL hyper-parameters live in \texttt{src/\bk{}FineTuning/\bk{}FunctionGemma/\bk{}finetune\_\bk{}config.yaml} (full SFT, 2 epochs, $\text{lr}=10^{-5}$, cosine, bf16, completion-only loss). The trainer is invoked through \texttt{main.py}; the resulting checkpoint is pushed back to Hugging Face via \texttt{to\_cloud.py}.
The source code is available 
\href{https://github.com/UgrowAI/Research_FuncGemma_FineTuning}{here}.

\paragraph{Step 2b: device export (\texttt{To\_Device.py}).} Convert a fine-tuned checkpoint to a \texttt{.litertlm} bundle for Handy Chat using \texttt{litert\_torch.generative.export\_hf.export} with \texttt{dynamic\_wi8\_afp32} quantization and \texttt{bundle\_litert\_lm=True} (Appendix~\ref{app:export}).

\paragraph{Step 3: evaluation (\texttt{\bk{}FGemma\_\bk{}Evaluation}).} \texttt{main.py} loads any number of model IDs (we used the four of Section~\ref{sec:experiments}) and any number of dataset IDs, runs greedy generation, parses outputs with the regex decoder of Appendix~\ref{app:parser}, computes the three metrics of Section~\ref{sec:task}, and writes per-row CSV/JSON, an aggregate pickle, and a free-text \texttt{review\_\bk{}result\_*.txt} for failure inspection. The same script produces all numbers in Tables~\ref{tab:main}, \ref{tab:percat-extended}, and \ref{tab:percat-google}.
The source code is accessible 
\href{https://github.com/UgrowAI/Research_FuncGemma_Evaluation}{here}.

\paragraph{Hugging Face artifact IDs.}
\begin{sloppypar}
\begin{itemize}[leftmargin=1.2em,itemsep=2pt,topsep=2pt]
    \item Datasets:
        \texttt{google/mobile-actions}~\citep{google-mobile-actions-dataset-2025},
        \dataExtRepo{},
        \dataMergeRepo{} (merged corpus).
    \item Models:
        \texttt{google/\bk{}functiongemma-270m-it}~\citep{functiongemma-2025},
        \texttt{litert-community/\bk{}FunctionGemma\_\bk{}270M\_\bk{}Mobile\_\bk{}Actions}~\citep{functiongemma-mobile-actions-2025},
        \modelExtRepo{},
        \modelCombRepo{},
        \modelExtLitRepo{},
        \modelCombLitRepo{}
        .
\end{itemize}
\end{sloppypar}

\subsection{FunctionGemma Output Parser}
\label{app:parser}

FunctionGemma encodes calls with sentinel tokens. The minimal parser used in our evaluation pipeline is reproduced below (Python).

\begin{lstlisting}[language=Python,style=acljson]
import re

def extract_function_call(model_output):
    """Parse <start_function_call>...<end_function_call> blocks.

    Each block has the form:
        call:func_name{key:<escape>value<escape>, key:<escape>value<escape>}
    Returns a list of {"function": {"name": ..., "arguments": ...}}.
    """
    results = []
    raw_calls = re.findall(
        r"<start_function_call>(.*?)"
        r"<end_function_call>",
        model_output,
        re.DOTALL,
    )
    for raw_call in raw_calls:
        if not raw_call.strip().startswith("call:"):
            continue
        try:
            pre_brace, args_segment = raw_call.split("{", 1)
            function_name = pre_brace.replace("call:", "").strip()
            args_content = args_segment.strip()
            if args_content.endswith("}"):
                args_content = args_content[:-1]
            arguments = {}
            for m in re.finditer(
                r"(?P<key>[^:,]*?):<escape>(?P<value>.*?)<escape>",
                args_content, re.DOTALL,
            ):
                arguments[m.group("key").strip()] = m.group("value")
            results.append({"function": {
                "name": function_name, "arguments": arguments}})
        except ValueError:
            continue
    return results
\end{lstlisting}

\subsection{Per-Category Counts of \datasetExtended{}}
\label{app:dataset-stats}

Eval-split sample counts per first-call category (multi-call examples are listed under their first call):

\begin{center}
\small
\begin{tabular}{lr}
\toprule
\textbf{Category} & \textbf{Eval $n$} \\
\midrule
\texttt{open\_application}      & 209 \\
\texttt{phone\_sms}             & 114 \\
\texttt{turnOn\_light}          &  73 \\
\texttt{bluetooth\_status}      &  70 \\
\texttt{take\_screenshot}       &  67 \\
\texttt{battery\_status}        &  65 \\
\texttt{list\_application}      &  56 \\
\texttt{get\_current\_brightness} & 53 \\
\texttt{turnOff\_light}         &  52 \\
\texttt{increase\_brightness}   &  45 \\
\texttt{phone\_call}            &  42 \\
\texttt{take\_picture}          &  31 \\
\texttt{decrease\_brightness}   &  31 \\
\texttt{send\_email}            &  29 \\
\texttt{set\_brightness}        &  28 \\
\midrule
\textbf{Total} & \textbf{965} \\
\bottomrule
\end{tabular}
\end{center}

\noindent The 90/10 train/eval split was applied at the record level via the dataset's \texttt{metadata} flag, giving a training set of $\sim$8{,}500 examples.

\subsection{Demo Traces}
\label{app:demo}

We list two representative end-to-end traces from Handy Chat (\modelOurs{} served via LiteRT-LM~\citep{litert-lm}); the \texttt{user} field is the typed prompt, the \texttt{predicted} field is the parsed tool call, and the \texttt{outcome} field is the Plyer~\citep{plyer} dispatch result.

\paragraph{Single call.}
\begin{demotrace}
user      : "Open Spotify and set the brightness to 80%."
predicted : [open_application(application_name="Spotify"),
             set_brightness(level="80")]
outcome   : OK -- Spotify launched, brightness set to 80
\end{demotrace}

\paragraph{Multi-call SMS + call.}
\begin{demotrace}
user      : "Text +14165550101 saying 'I am on my way.' and then
             call the same number directly."
predicted : [phone_sms(sms_recipient="+14165550101",
                       sms_message="I am on my way."),
             phone_call(number="+14165550101")]
outcome   : OK (SMS dispatched), then phone call placed.
note      : argument metric flags this as IMPERFECT because the gold
            label includes "dial_call: false"; the user-visible
            outcome is correct.
\end{demotrace}

\subsection{Dataset Format Reference}
\label{app:format}

Each record in \datasetExtended{} (and \datasetGoogle{}) is a JSON object with three top-level fields: \texttt{metadata}, \texttt{tools}, and \texttt{messages}. \texttt{metadata} is the literal string \texttt{train} or \texttt{eval}. \texttt{tools} is the catalog of Appendix~\ref{app:schema}. \texttt{messages} is an ordered list of role-tagged turns:

\begin{itemize}[leftmargin=1.2em,itemsep=2pt,topsep=2pt]
    \item \texttt{developer}: ``You are a model that can do function calling with the following functions \ldots'', followed by the current date/time/weekday and the function declarations rendered with the FunctionGemma sentinel tokens.
    \item \texttt{user}: the natural-language utterance.
    \item \texttt{assistant}: the gold \texttt{tool\_calls} list of \texttt{\{"function": \{"name": \ldots, "arguments": \{\ldots\}\}\}} objects.
\end{itemize}

\noindent At training time, the chat template renders this list into a single string; at evaluation time, our parser of Appendix~\ref{app:parser} converts the model's continuation back into \emph{(name, args)} tuples.


\section{Handy Chat: On-Device Demo Application}
\label{app:demo-section}

We deploy .litertlm version of \modelOurs{} in \emph{\href{https://github.com/UgrowAI/HandyChat-apk/releases/latest/download/HandyChat.apk}{Handy Chat}}, a standalone Android APK built with Buildozer~\citep{buildozer} (\texttt{org.earthisgreen.funcgemmachat}). The Google AI Edge Gallery~\citep{ai-edge-gallery} does not currently allow loading custom checkpoints into its pre-configured Mobile Actions workflow; we therefore ship our own runtime.

\subsection{Architecture}
\label{app:demo-arch}

The app combines four layers:

\begin{enumerate}[leftmargin=1.4em,itemsep=2pt,topsep=2pt]
    \item \textbf{Kivy UI (typed input only).} A Kivy/KivyMD chat tab displays conversation history and a text field with a Send button. \emph{Speech input is not enabled} in the current release (the microphone control is unused).
    \item \textbf{LiteRT-LM via Java bridge (CPU).} On-device inference uses the \texttt{litertlm-android} library with the \textbf{CPU} backend (\texttt{Backend.CPU}). A thin Java class (\texttt{litertlm\_kotlin}) wraps \texttt{Engine.initialize()}, \texttt{createConversation()}, and \texttt{sendMessage()}; Python invokes it through PyJNIus on the Android UI thread.
    \item \textbf{Conditional two-pass loop.} The Python layer builds FunctionGemma chat-template prompts and sends the user turn to LiteRT-LM (\textbf{pass~1}). If the regex parser (Appendix~\ref{app:parser}) detects a function call, each call is executed through \texttt{FunctionExecutor}/Plyer and serialized tool responses are sent back for \textbf{pass~2}; if no function call is found, the app returns the pass~1 text directly without a second decode.
    \item \textbf{Plyer dispatch.} Registered tool handlers mirror the fifteen extended categories and invoke Android intents (\texttt{turnOn\_light}, \texttt{phone\_sms}, \texttt{open\_application}, etc.) via Plyer~\citep{plyer}.
\end{enumerate}

\subsection{Checkpoint Export to \texorpdfstring{\texttt{.litertlm}}{.litertlm}}
\label{app:export}

Handy Chat consumes a LiteRT-LM bundle rather than the raw Hugging Face weights. We export offline with \texttt{To\_Device.py} in our fine-tuning repository. The script loads the fine-tuned checkpoint directory and calls
\begin{quote}\small\texttt{litert\_torch.generative.export\_hf.export(}\\
\texttt{\ \ model=checkpoint\_dir,}\\
\texttt{\ \ quantization\_recipe="dynamic\_wi8\_afp32",}\\
\texttt{\ \ output\_dir=..., bundle\_litert\_lm=True)}
\end{quote}
\noindent which produces a \href{https://huggingface.co/UGrowAI/FunctionGemma-270M-it-Mobile-Actions-Extended-litert-lm/tree/main}{\texttt{.litertlm} file}\modelOursLitRepo (285.6\,MB for our extended checkpoint) bundled with the tokenizer artifacts needed by the on-device runtime. Export runs on a development machine; the APK ships the bundle under application assets.

The on-device loop is: typed message $\to$ pass~1 decode $\to$ parse. When a function call is present, Plyer dispatches it and pass~2 ingests the tool response; otherwise pass~1 output is shown to the user. We verified all fifteen extended categories on a developer device. Representative traces are in Appendix~\ref{app:demo}. The demo APK and source are released alongside the Hugging Face artifacts.

\subsection{Hardware, Device, and Latency Details}
\label{app:demo-hw}

All demo measurements were taken on a \textbf{Google Pixel 8 Pro} (``husky''): Tensor~G3 SoC, 12\,GB RAM, Android~16. The Mali-G715 GPU is present but \emph{not} used for inference; Handy Chat runs entirely on the \textbf{LiteRT-LM CPU backend} (\texttt{Backend.CPU}). The deployed model is the 285.6\,MB \texttt{.litertlm} bundle exported with the \texttt{dynamic\_wi8\_afp32} recipe (Appendix~\ref{app:export}); note that this dynamic INT8 quantization affects only the demo, and all accuracy numbers in the paper were measured on the 
unquantized 
base or fine-tuned models.

Across a handful of prompt trials, a function-call turn takes roughly \textbf{13 seconds average end-to-end latency}, broken down by stage as follows: pass~1 (prompt prefill and tool-call decode) $\sim$9\,s; tool execution through Plyer $\sim$2\,s; pass~2 (ingesting the serialized tool response and producing the user-facing reply) $\sim$2\,s. Turns without a function call skip stages two and three and return the pass~1 output directly.

\end{document}